# Comparison of deep learning models: CNN and VGG-16 in identifying pornographic content

Reza Chandra[1], Adang Suhendra[2], Lintang Yuniar Banowosari[2], Prihandoko[3]
[1]Department of Informatics Management, Gunadarma University, Depok City, Indonesia
[2]Department of Informatics, Gunadarma University, Depok City, Indonesia
[3]Department of Information Systems, Gunadarma University, Depok City, Indonesia

**ABSTRACT**

In 2020, a total of 59,741 websites were blocked by the Indonesian government due to containing negative content, including pornography, with 14,266 websites falling into this category. However, these blocked websites could still be accessed by the public using virtual private networks (VPNs). This prompted the research idea to quickly identify pornographic content. This study aims to develop a system capable of identifying websites suspected of containing pornographic image content, using a deep learning approach with convolutional neural network (CNN) and visual geometry group 16 (VGG-16) model. The two models were then explored comprehensively and holistically to determine which model was most effective in detecting pornographic content quickly. Based on the findings of the comparison between testing the CNN and VGG-16 models, research results showed that the best test results were obtained in the eighth experiment using the CNN model at an epoch value level of 50 and a learning rate of 0.001 of 0.9487 or 94.87%. This can be interpreted that the CNN model is more effective in detecting pornographic content quickly and accurately compared to using the VGG-16 model.

*Corresponding Author:*

Reza Chandra
Department of Informatics Management, Gunadarma University
Margonda Raya 100 Road, Depok City 16424, Indonesia
Email: reza_chan@staff.gunadarma.ac.id

## 1. INTRODUCTION

Currently, more and more Indonesian people are acquiring cultural values and norms that are not relevant to moral values. It is still common to find television shows, including film material, which often show pornographic scenes that are not in accordance with the values and first principles of the state principles of Pancasila. In the midst of current advances in digitalization, the culture of media platforms must adapt to society's needs. Television and film media, including smartphone providers and developers, have a strategic role in disseminating information.

Problems arise when providers of television broadcasts, films, and social media grow freely without any intervention from the government. By prioritizing the principle of freedom of expression, it is feared that society could be divided into public opinion as part of a form of attitude in the form of fighting against one another and this could easily lead to horizontal escalation [1]. The Indonesian government, through the leadership of President Joko Widodo, continues to make efforts with all its policy instruments to block various websites deemed and suspected to have negative and provocative content that can protect the positive thoughts of the public, to create a safe, positive and high-quality internet atmosphere, one of which is pornographic content [2], [3].

Web usage is growing in a way that is not well controlled. An appropriate framework model is needed to limit the vast amount of information in cyberspace. With the advent of the internet, pornography became cheap, many porn sites could be easily built and accessed. Pornography is still considered illegal and against the law in Indonesia, thus encouraging the government's active involvement in developing regulations to limit it [4]–[6].

The researcher's observations are that until now the Indonesian government can only carry out blocking actions based on web addresses and not based on image content. Based on the web page https://aduankonten.id owned by The Ministry of Communication and Information (Kominfo), as of March 2020, there were 1,246,801 public complaints regarding negative content reported via this page. Web monitoring is carried out based on 21 categories: categories of hoaxes, slander, intellectual property rights (IPR), violence/violence against children, content that violates social and cultural values, information security violations, extortion, fraud, trade in products with special regulations, gambling, pornography, ethnicity, religion, race and class (SARA), separatism, and terrorism. Pornography websites received the most complaints from the public with 1,037,321 complaints as shown in Figure 1.

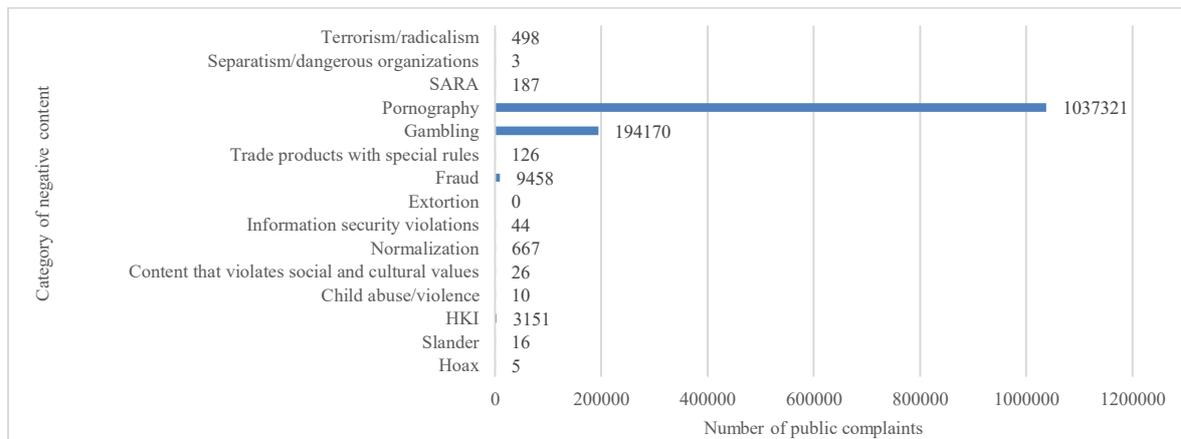

Figure 1. Statistics on public complaints regarding negative content

Given the increasing exposure to pornography in Indonesia, the potential negative psychological impacts of exposure to pornography (especially those related to potential problems or addiction) have become an increasing empirical concern in recent years [7]. Hu *et al.* [8] carried out pornographic image recognition where the current methods can be classified as model, feature, and region based. They used skin detection for pornographic image recognition. The weakness in the research, preprocessing is required in classifying images such as skin color extraction and the problems of over blocking, misspellings and word lists have not yet been resolved. convolutional neural network (CNN) is the model most often used in image processing and computer vision [9].

The using CNN models to learn high-level features in image data through convolution. CNN is very suitable for object recognition with images. The superiority of CNN in image recognition is the main argument as well as the world's recognition of the superiority of deep learning in studying large amounts of data and its ability in remote sensing to achieve advanced classification accuracy [10]–[12]. Even though this model is very good at classifying certain images, the CNN model is in fact still not optimally used as a filter for content that is indicated to have pornographic elements by many researchers, most of whose solutions are still cloud-based [13]–[15].

The researcher's perspective uses the CNN model for more accurate recognition of pornographic image content. The CNN model is built with three convolution layers. The amount of data used is 14,085 image data. Researchers also use the visual geometry group 16 (VGG-16) model as a comparison model, this model is a development of the CNN model which has 16 convolution layers. Agastya *et al.* [13] used the CNN model for classifying pornographic images and developed the idea in this research. Researchers propose a strategy for implementing web image content identification using a deep learning approach (CNN and VGG-16) as a filter for websites that contain pornographic image content. The novelty built in this research is a deep learning approach. The advantage obtained in this research is that it can identify provocative pornographic image content in cyberspace quickly, helping the government provide safe and quality internet protection for the public in accessing websites that are detected to contain pornographic content, so this research does not require preprocessing in classifying image data.

## 2. THE COMPREHENSIVE THEORETICAL BASIS
### 2.1. Deep learning based on artificial neural networks

Through iterative processing and training on large datasets, deep learning models can automatically extract and transform features, making them highly effective for complex tasks such as image recognition, natural language processing and speech analysis. Deep learning is the multi-layer artificial neural networks that can learn the complex non-linear functions. Deep learning refers to data-driven learning that uses multi-layer neural networks and hierarchical processing to compute complex patterns and representations from large amounts of data [16]. This theory is a form of machine learning process that is based on an artificial neural network with many hidden layers and its depth in analysing representations or data features automatically which is included in the category of artificial intelligence [17], [18]. This theory is a machine learning process with the characteristics of using several layers of hierarchical non-linear information processing, this is determined by how the intended architecture and techniques are used, for example synthesis or classification. Deep learning is divided into 3 main class categories [19]: i) unsupervised or generative deep network learning. In the latest case, the use of Bayes' rules can change the characteristics of a generative network into a discriminative network for learning; ii) deep network supervised learning provides discriminative capabilities for patterns by clustering the posterior distribution of groups conditioned on the visible purposes data; and iii) hybrid deep networks (assisted discrimination), often in a significant way and the result of generative deep networks or unsupervised learning. Maximizing the form of network regularization in the supervised learning category. Goodfellow *et al.* [20] argued deep learning uses hierarchical concepts to solve problems in computer learning systems.

### 2.2. Image processing based on convolutional neural network model

The CNN model is the most frequently used model for image processing and computer vision [21]. This model is designed in such a way as to mimic the structure of the visual cortex. Specifically, a CNN has the form of neurons in a particular layer only connecting to a small area of the previous layer. This model is most often used for image processing and computer vision. This model included in the form of a deep neural network, this is because of its characteristics and high network depth and its use is often applied to image data that is being observed [22]. The classification results are produced from the output layer where each part of the hyperspectral image is directed to a model that indicates back and forth propagation. Hyperspectral image classification is a crucial step in hyperspectral data processing [23]. Highly consistent results by combining finger veins and facial feature through an iterative premium local extraction process with this model [24]. The results of the latest research regarding the use of this model in a biometric framework through combining finger veins and facial features as well as bimodal layer fusion and utilizing the AlexNet as well as VGG-16 models produce high efficacy in differentiating finger veins images from facial features. The results of both models have authentication accuracy reaching 98.4% [25], [26].

### 2.3. Classifying images with the VGG-16 model

We also use the VGG-16 model to represent the CNN structure using 13 convolution layers, 5 max pools, 1 flatten layer, and 3 dense layers. This model utilizes a compact 3×3 convolutional filter as well as a deep architecture using size 1. While the padding layer adopts a 2×2 configuration at a step size of 2 while maintaining equal padding [27]. VGG-16 model performs 224×224 input image processing. Before the fully connected layer, a 7×7 feature map containing 512 channels is used. Then, the results of the feature mapping are processed into a vector with 25,088 channels (7×7×512) as a representation of the resulting features [28], [29]. Apart from the CNN model, this research also uses the VGG-16 model as a comparison to see which model is better at identifying pornographic content with data of 14,085 images.

### 2.4. Level of access and regulation regarding pornography in Indonesia

In Indonesia, there is a condition where quite a lot of people are interested in pornography via the internet. Along with communication and information technology, the spread of pornographic media can be accessed online via the internet. Internet pornography has not only developed as a personal need, but has also become a commodity that is sold commercially and professionally. Pornographic news is something that can be sold in the mass media, apart from news about crime such as cyber-sex and porn sites. Pornography is classified as a criminal act that violates morality (Zedelijkheid) and is a problem related to sexuality. Pornography in Indonesia is considered illegal and violates the law [4]. Pornography is regulated in article 1 number (1) in regulation law 44/2008 concerning pornography and the information and electronic transactions (*ITE*) law chapter VII concerning prohibited acts, especially in article 27 paragraph (1) and law number 36 1999 paragraph 21 concerning telecommunications, especially points regarding morality (pornography).

## 3. METHOD
### 3.1. Research objects
Websites that were allegedly detected to contain pornographic content were the objects of this research. This website was taken based on article information from the official Kominfo website of the Republic of Indonesia on the website https://trustpositive.kominfo.go.id/ which can be accessed by the public. There are 773,517 websites that are blocked, but there are several websites that can still be accessed using the virtual private network (VPN) application and there are also those whose domain names have expired.

### 3.2. Research framework
One of the reasons for using the CNN model for image identification is because this model has the most significant results in recognizing image objects. CNN can recognize images with almost the same level of accuracy as humans on a certain number of datasets [30]. Algorithm and construction of a pornographic image content identification system: i) hardware design, using a DGX A100 machine equipped with 8 GPUs and having a total memory of 320 GB and equipped with 6 NVIDIA NVSwitches because the matrix operation computation in CNN and VGG-16 is very intensive. GPUs designed to perform parallel computations are well suited to this task; ii) software system, using TensorFlow and Keras as a deep learning framework to build and train CNN and VGG-16 models. The programming language uses Python because of its rich ecosystem and the large number of libraries available. Supporting libraries use OpenCV for image processing, NumPy (numerical computing), Pandas (frame data), Matplotlib (data visualization); and iii) the HTTP protocol is used to send requests and receive responses from the server when scraping data from websites.

### 3.3. Convolutional neural network model design
Image data collection based on references to websites indicated to contain pornographic content which are blacklisted by Kominfo via the web trustpositive.kominfo.go.id and a collection of websites containing non-pornographic content from various sources. List of website addresses containing pornographic content as shown in Table 1. To obtain an image dataset, it is necessary to design a web crawler for websites that are deemed to have pornographic and non-pornographic content. The stages are as in Figure 2.

The first step, choose websites that are indicated as pornographic and non-pornographic. Then enter the website URL and carry out a crawling process which will pull all the images on the targeted website and save them in a directory. When the data has finished crawling, the data is grouped into two directories, namely the test set directory which is the directory for the test image dataset which consists of pornographic image datasets and non-pornographic image datasets and the training set directory which is the directory for the training image dataset which consists of pornographic image datasets and non-pornography, (Figure 3). After successfully obtaining a dataset of pornographic images and a dataset of non-pornographic images, we then designed a CNN process to train the dataset and carry out its validity with test data using 14,085 data with a training data composition of 80%, the remaining test data is 20%. The initial stage of designing the CNN architectural model starts from the input data, then goes to the 2D zero padding layer which is used to become the output size whose shape and proportions are the same as the input size. This model consists of 3 2D convolutional layers, 3 2D max pool layers, 1 flatten layer and 2 dense layers as in Figure 4 and then identifies image content through several stages as in Figure 5.

Table 1. Web address indicated to contain pornographic content

| No | Website URL | IP address | Access time | Crawling time |
|---|---|---|---|---|
| 1 | asiannewpics.com/ | 69.10.41.18 | 04/10/2019 | 3 hours |
| 2 | sexpics24.com/ | 88.85.75.41 | 03/28/2019 | 3 hours |
| 3 | asianbabepics.com/ | 109.232.227.100 | 04/10/2019 | 3 hours |
| 4 | nudesexporn.com/ | 23.226.129.188 | 04/10/2019 | 4 hours |
| 5 | pornpics.com/ | 108.60.221.55 | 03/28/2019 | 4 hours |
| 6 | asianpornpicture.com/ | 109.232.227.100 | 03/28/2019 | 3 hours |
| 7 | highasianporn.com/ | 162.251.111.236 | 03/28/2019 | 2 hours |
| 8 | allbabesnaked.com/ | 185.73.221.118 | 03/28/2019 | 2 hours |
| 9 | chinesemifl.com/ | 109.232.227.101 | 04/10/2019 | 2 hours |
| 10 | lamalinks.com/ | 208.69.117.6 | 03/28/2019 | 3 hours |
| 11 | hdpornpics.com/ | 206.54.175.4 | 04/10/2019 | 4 hours |
| 12 | forcesexpic.com/ | 88.76.59.10 | 04/10/2019 | 1 hour |
| 13 | nudexxxpictures.com/ | 78.140.153.205 | 03/28/2019 | 2 hours |
| 14 | pornpicsamateur.com/ | 78.140.133.54 | 04/10/2019 | 2 hours |
| 15 | nudeblackgirlsphotos.com/ | 64.111.197.114 | 03/28/2019 | 2 hours |
| 16 | pornpics.io/ | 88.85.75.40 | 04/10/2019 | 1 hour |
| 17 | youxx.xx.com/ | 185.53.178.9 | 03/28/2019 | 2 hours |
| 18 | wowerotica.com/ | 199.80.52.155 | 04/10/2019 | 4 hours |
| 19 | babesource.com/ | 99.192.136.176 | 03/28/2019 | 3 hours |
| 20 | dirtypornphotos.com/ | 88.85.75.37 | 04/10/2019 | 3 hours |

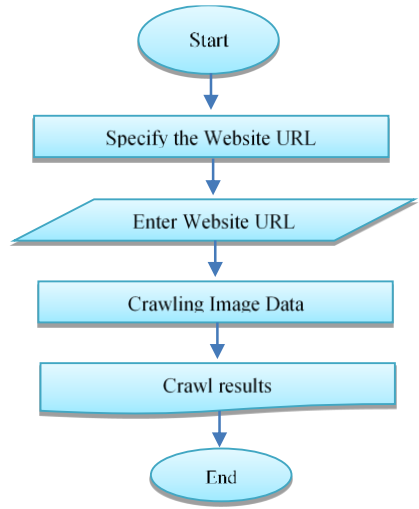

Figure 2. Image content crawling flowchart

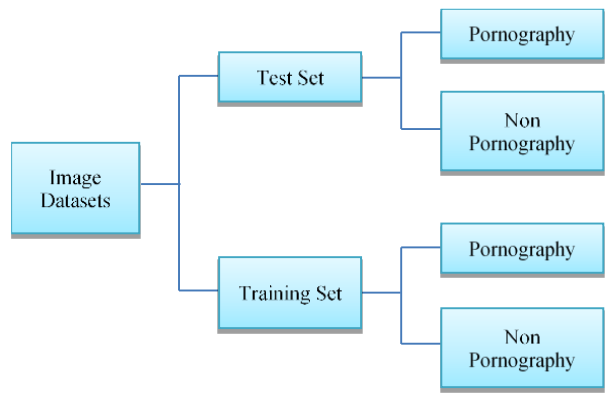

Figure 3. Image data labeling design

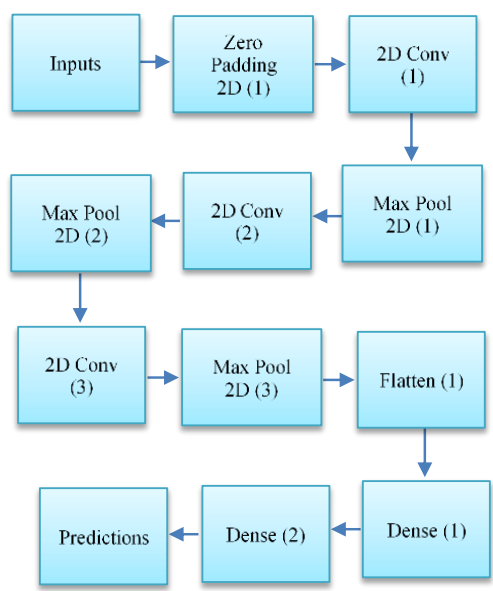

Figure 4. CNN model architecture

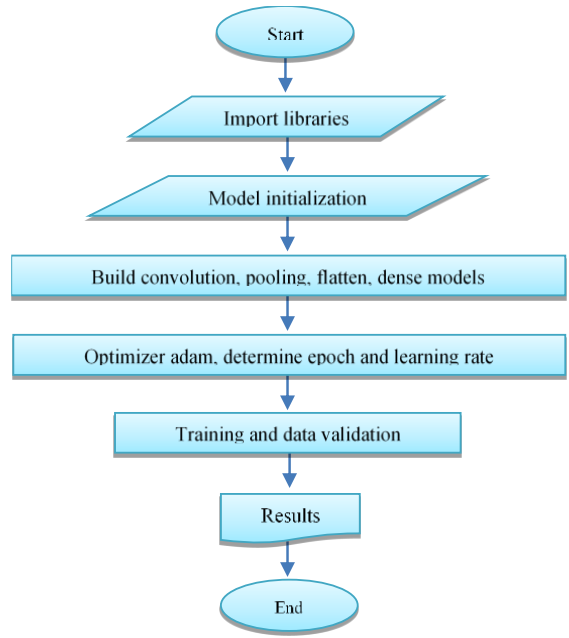

Figure 5. CNN model flowchart for image identification

Entering the libraries needed to start the process. Create a zeropadding layer which is the first layer needed in the process, then the first convolutional layer needed after that the process continues with max pooling 2D and then goes to the second convolutional layer then the final max pooling 2D process before the layer is finished being created. Determine the algorithm used for the optimizer, namely Adam, determines the number of epochs and determines the learning rate value; start the data training process; save the results of the training data in the form of an h5 model; and the CNN process is complete.

### 3.4. VGG-16 model design

The design of this model aims to compare the image identification results from the CNN model. This model uses 13 convolution layers, 5 max pool layers, 1 flatten layer and 3 dense layers. The flow is the same as the CNN model because the VGG-16 model is a derivative of the CNN model, see Figure 6.

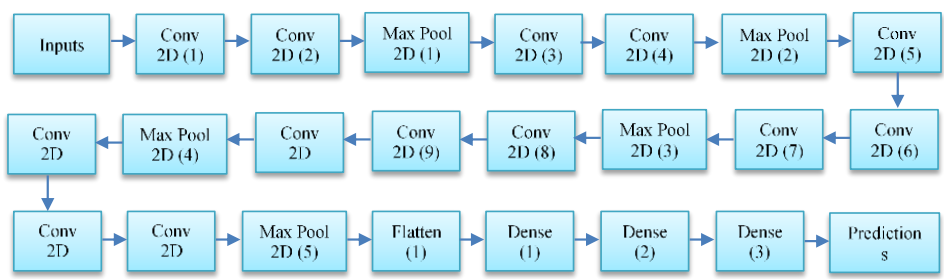

Figure 6. Flow architecture in the VGG-16 model

## 4. RESULTS AND DISCUSSION
### 4.1. Identification of image content identified as pornographic with the CNN model

To crawl images, a multithreading method is used through several steps including: i) determining websites that are indicated as pornographic and non-pornographic, namely determining the destination web address, and retrieving web page information; ii) carry out the crawling process starting using the multithreading method on the website by retrieving image data on the website via for links in links, inputting image links, and extracting image data; and iii) the results of the crawling process were successfully saved. To train and test the image dataset, the CNN algorithm is used. This algorithm was compiled using the Python programming language model version 3.6 and the Keras library. The steps for developing a CNN model are as follows:

i) inserting the NumPy, TensorFlow, TensorBoard and Keras libraries; ii) CNN initialization; iii) building the CNN model includes building the first layer by determining padding, activation function and max pooling; build the second layer by determining the activation and max pooling functions; and build the third layer by determining the activation and max pooling functions; iv) add flatten; v) add 2 fully connected layers; vi) add optimizer; vii) call the TensorBoard library for graphics; viii) fitting process; ix) training process; and x) save the training results. In testing the identification of image content that is indicative of pornography through CNN model training, 9 trials were carried out using different epoch variable values, different activation functions, different learning rates, so that the best results were obtained from the model that had been built, as follows.

### 4.1.1. First test results of convolutional neural network model

Based on the parameters: i) number of layers=1; ii) learning rate=0.001; iii) epoch=25; iv) steps per epoch=100; and v) activation function at the full connection layer=rectified linear units (ReLu) and sigmoid. Using the parameter number of convolution layers is 1, the learning rate has a value of 0.001, epoch has a value of 25, steps per epoch has a value of 100 and selecting the ReLu activation function in the full connection layer produces an accuracy value of 0.8341, the loss value reaches 0.3691, valid accuracy has a value of 0.8299 and a valid loss value of 0.2804. Figure 7 shows first results of identification of image content identified as pornographic with the CNN model.

### 4.1.2. Second test results convolutional neural network model

Based on the parameters: i) number of layers=1; ii) learning rate=0.0001; iii) epoch=25; iv) steps per epoch=100; and v) activation function at the full connection layer=ReLu and sigmoid. Used parameters for the number of convolution layers of 1, the learning rate was 0.0001, the epoch was 25, the steps per epoch was 100 and the ReLu activation function on the full connection layer produced an accuracy value of 0.8228, a loss value of 0.386, a valid value of accuracy is 0.8038 and a valid loss value is 0.4707. Figure 8 shows second results of identification of image content identified as pornographic with the CNN model.

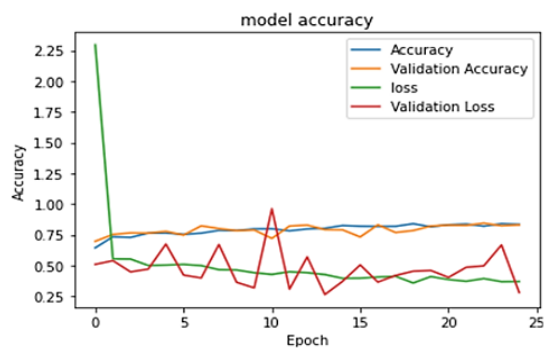
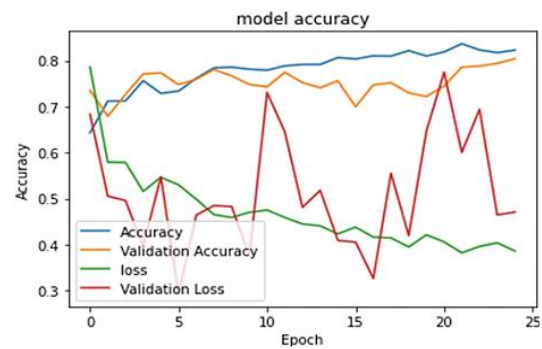

Figure 7. First results of identification of image content identified as pornographic with the CNN model

Figure 8. Second results of identification of image content identified as pornographic with the CNN model

### 4.1.3. Third test results convolutional neural network model

Based on the parameters: i) number of convolution layers=1; ii) learning rate=0.001; iii) epochs=50; iv) steps per epoch=100; and v) activation function at the full connection layer=Relu and sigmoid. Used the parameters of the number of convolution layers being 1, the learning rate is 0.001, the epoch is 25, the steps per epoch are 100 and the Relu activation function on the full connection layer produces an accuracy value of 0.8744, a loss value of 0.2838, a valid value. accuracy is 0.8289 and valid loss value is 0.2281. Figure 9 shows third results of identification of image content identified as pornographic with the CNN model.

### 4.1.4. Fourth test results convolutional neural network model

Based on the parameters: i) number of convolution layers=2; ii) learning rate=0.0001; iii) epoch=25; iv) steps per epoch=100; and v) activation function at the full connection layer=ReLu and sigmoid. Using the parameter number of convolution layers is 2, the learning rate is 0.0001, the epoch is 25, the steps per epoch is 100 and the ReLu activation function on the full connection layer produces an accuracy value of 0.8503, the loss rate value reaches 0.3475, valid accuracy has a value of 0.7509 and a valid loss value of 0.2809. Figure 10 shows fourth results of identification of image content identified as pornographic with the CNN model.

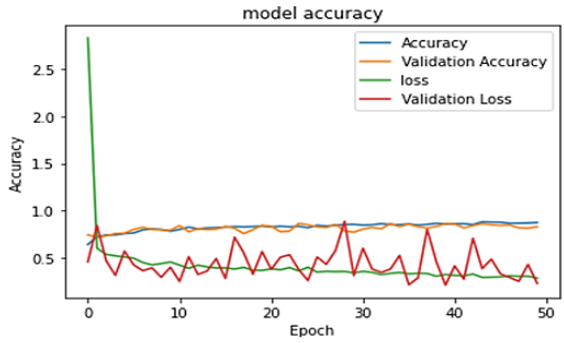

Figure 9. Third results of identification of image content identified as pornographic with the CNN model

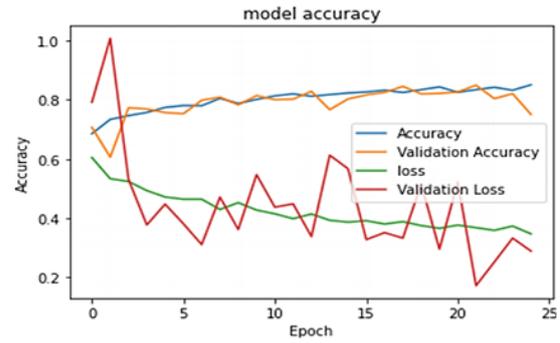

Figure 10. Fourth results of identification of image content identified as pornographic with the CNN model

**4.1.5. Fifth test results convolutional neural network model**

Based on the parameters: i) number of convolution layers=2; ii) learning rate=0.001; iii) epochs=50; iv) steps per epoch=100; and v) activation function at the full connection layer=ReLu and sigmoid. Using the parameter number of convolution layers is 2, the learning rate has a value of 0.001, epoch has a value of 50, steps per epoch has a value of 100 and selecting the ReLu activation function in the full connection layer produces an accuracy value of 0.9066, the loss value reaches 0.218, a valid value of accuracy of 0.8435 and the valid loss value is 0.2342. Figure 11 shows fifth results of identification of image content identified as pornographic with the CNN model.

**4.1.6. Sixth test results convolutional neural network model**

Based on the parameters: i) number of convolution layers=2; ii) learning rate=0.0001; iii) epochs=50; iv) steps per epoch=100; and v) activation function at the full connection layer=ReLu and sigmoid. Using the parameter number of convolution layers is 2, the learning rate is 0.0001, the epoch is 50, the steps per epoch is 100 and the ReLu activation function on the full connection layer produces an accuracy value of 0.8741, while the loss value reaches 0.3001, valid accuracy has a value of 0.7673 and a valid loss value of 0.4083. Figure 12 shows sixth results of identification of image content identified as pornographic with the CNN model.

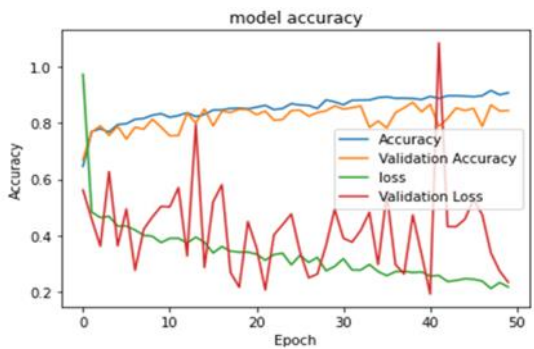

Figure 11. Fifth results of identification of image content identified as pornographic with the CNN model

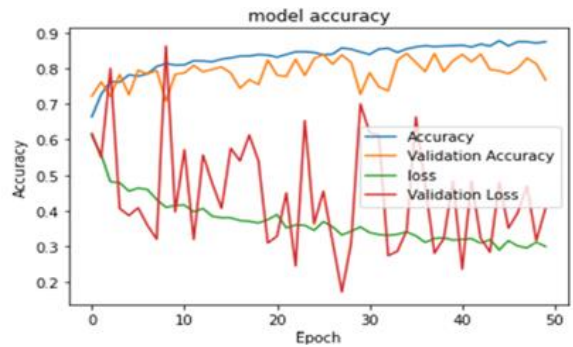

Figure 12. Sixth results of identification of image content identified as pornographic with the CNN model

**4.1.7. Seventh test results convolutional neural network model**

Based on the parameters: i) number of convolution layers=3; ii) learning rate=0.0001; iii) epoch=25; iv) steps per epoch=100; and v) activation function at the full connection layer=ReLu and sigmoid. Using the parameter number of convolution layers is 3, the learning rate is 0.0001, the epoch is 25, the steps per epoch is 100 and the ReLu activation function on the full connection layer produces an accuracy value of 0.8331, the loss value is 0.3752, valid accuracy has the value is 0.8254, and the valid loss value is 0.6289. Figure 13 shows seventh results of identification of image content identified as pornographic with the CNN model.

### 4.1.8. Eighth test results convolutional neural network model

Based on the parameters: i) number of convolution layers=3; ii) learning rate=0.001; iii) epochs=50; iv) steps per epoch=100; and v) activation function at the full connection layer=ReLu and sigmoid. Using the parameter number of convolution layers is 3, the learning rate has a value of 0.001, epoch has a value of 50, steps per epoch has a value of 100 and selecting the ReLu activation function in the full connection layer produces an accuracy value of 0.9487, loss reaches a value of 0.1787, valid accuracy is 0.8831, and valid loss value is 0.6308. Figure 14 shows eighth results of identification of image content identified as pornographic with the CNN model.

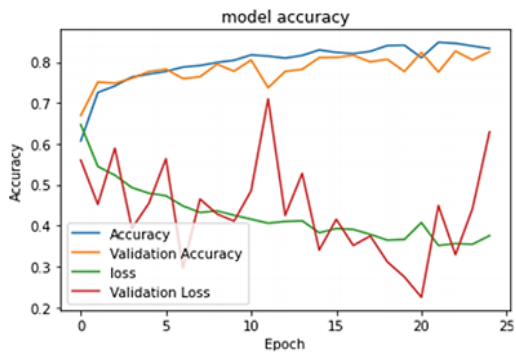
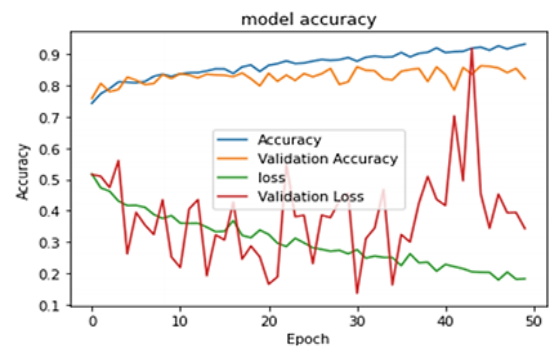

Figure 13. Seventh results of identification of image content identified as pornographic with the CNN model

Figure 14. Eighth results of identification of image content identified as pornographic with the CNN model

### 4.1.9. Ninth test results convolutional neural network model

Based on the parameters: i) number of convolution layers=3; ii) learning rate=0.001; iii) epoch=25; iv) steps per epoch=100; and v) activation function at the full connection layer=ReLu and sigmoid. Using the parameters of the number of convolution layers totaling 3, learning rate with a value of 0.001, epoch with a value of 25, steps per epoch with a value of 100 and the ReLu activation function on the full connection layer produces an accuracy value of 0.8122, the loss value reaches 0.417, the valid accuracy value reaches 0.7914, and valid loss has a value of 0.561. Figure 15 shows ninth results of identification of image content identified as pornographic with the CNN model.

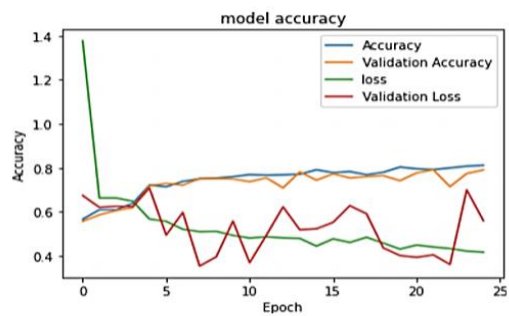

Figure 15. Ninth results of identification of image content identified as pornographic with the CNN model

### 4.2. Results of identification of image content identified as pornographic with the VGG-16 model

VGG model training is carried out by testing using different epoch variable values, different activation functions, different learning rates, so that the best results are obtained from the model that has been built, as follows:

### 4.2.1. First test results of VGG-16 model

The trial was carried out based on the parameters: i) learning rate=0.001; ii) epoch=100; iii) steps per epoch=100; and iv) activation function at the full connection layer=ReLu, ReLu, and softmax. Using the learning rate parameters of 0.001, epoch of 100, steps per epoch of 100 and the ReLu, ReLu, and softmax activation functions on the full connection layer produces an accuracy value of 0.5347, a valid accuracy value of 0.4375, a loss value of 0.6909, and the valid loss value also reached 0.7083. Figure 16 shows first results of identification of image content identified as pornographic with the VGG-16 model.

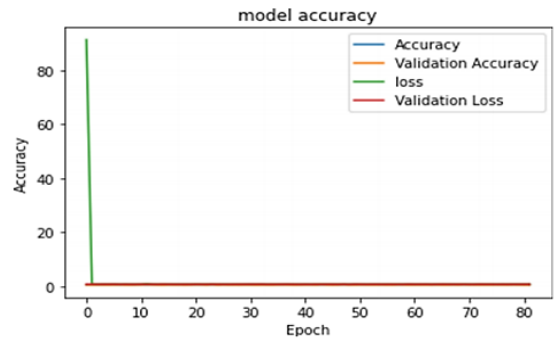

Figure 16. First results of identification of image content identified as pornographic with the VGG-16 model

### 4.2.2. Second test results

The trial was carried out based on the parameters: i) learning rate=0.001; ii) epochs=50; iii) steps per epoch=30; and iv) activation function at the full connection layer=ReLu, ReLu, and sigmoid. Using the learning rate parameters were 0.001, epoch was 50, steps per epoch were 30 and the ReLu, ReLu, and sigmoid activation functions in the full connection layer produced an accuracy value of 0.5198, a valid accuracy value of 0.4844, a loss value of 7.74, and a valid loss value of 7.0517. Figure 17 shows second results of identification of image content identified as pornographic with the VGG-16 model.

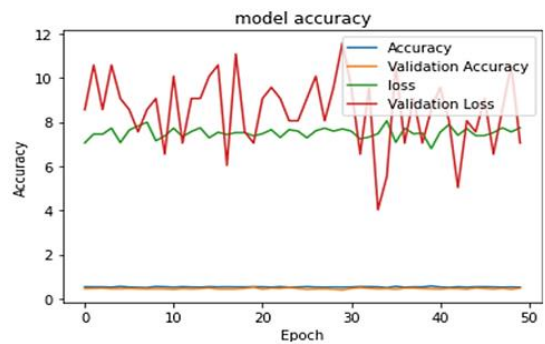

Figure 17. Second results of identification of image content identified as pornographic with the VGG-16 model

### 4.2.3. Third test results

The trial was carried out based on the parameters: i) learning rate=0.001; ii) epochs=50; iii) steps per epoch=50; and iv) activation function at the full connection layer=ReLu, ReLu, and ReLu. Using the learning rate parameters were 0.001, epoch was 50, steps per epoch were 30 and the ReLu, ReLu, and ReLu activation functions on the full connection layer produced an accuracy value of 0.4512, a valid accuracy value of 0.5188, a loss value of 8.8448, and a valid loss value of 7.5554. Figure 18 shows third results of identification of image content identified as pornographic with the VGG-16 model.

### 4.2.4. Fourth test results

The trial was carried out based on the parameters: i) learning rate=0.0001; ii) epochs=50; iii) steps per epoch=50; and iv) activation function at the full connection layer=ReLu, ReLu, and ReLu. Using the learning rate parameters were 0.0001, epoch was 50, steps per epoch were 50 and the ReLu, ReLu, and ReLu activation functions in the full connection layer produced an accuracy value of 0.4674, a valid accuracy value of 0.5437, the loss value was 1.3211, and the valid loss value is 3.5258. Figure 19 shows fourth results of identification of image content identified as pornographic with the VGG-16 model.

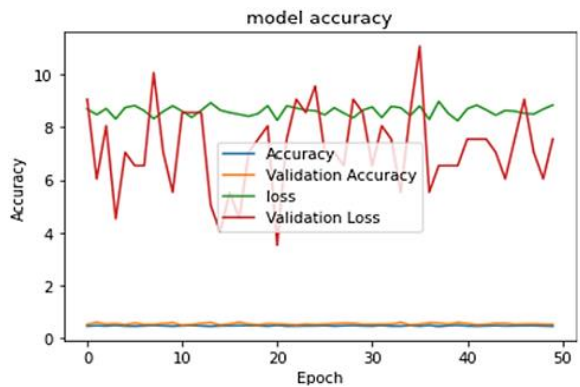

Figure 18. Third results of identification of image content identified as pornographic with the VGG-16 model

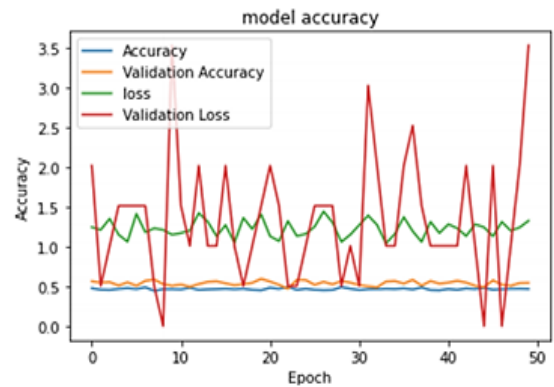

Figure 19. Fourth results of identification of image content identified as pornographic with the VGG-16 model

### 4.2.5. Fifth test results

The trial was carried out based on the parameters: i) learning rate=0.0001; ii) epochs=50; iii) steps per epoch=50; and iv) activation function at the full connection layer=softmax, softmax, and softmax. Used learning rate parameters of 0.0001, epoch of 50, steps per epoch of 50 and the softmax, softmax, and softmax activation functions on the full connection layer produced an accuracy value of 0.5344, a valid accuracy value of 0.5063, the loss value is 0.6908, and the valid loss value is 0.6994. Figure 20 shows fifth results of identification of image content identified as pornographic with the VGG-16 model.

### 4.2.6. Sixth test results

The trial was carried out based on the parameters: i) learning rate=0.0001; ii) epoch=30; iii) steps per epoch=100; and iv) activation function at the full connection layer=ReLu, ReLu, and softmax. Used learning rate parameters of 0.0001, epoch of 30, steps per epoch of 100 and the ReLu, ReLu, and softmax activation functions on the full connection layer produced an accuracy value of 0.9531, a valid accuracy value of 0.8125, the loss value is 0.1214, and the valid loss value is 0.3222. Figure 21 shows sixth results of identification of image content identified as pornographic with the VGG-16 model.

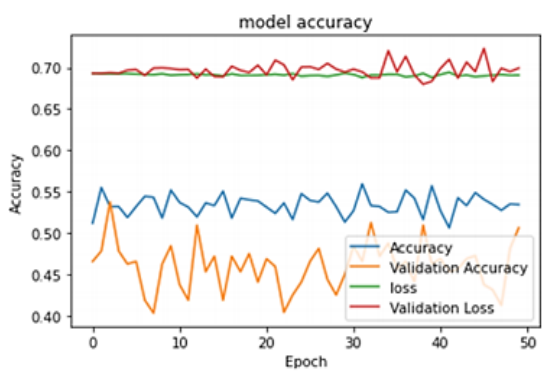

Figure 20. Fifth results of identification of image content identified as pornographic with the VGG-16 model

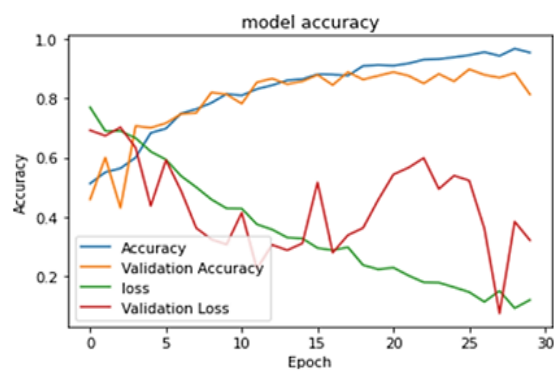

Figure 21. Sixth results of identification of image content identified as pornographic with the VGG-16 model

### 4.3. Comparison identification of pornographic image content with CNN and VGG-16 models

The following is a comparison of the identification of image content identified as pornographic using the two models, where testing with the CNN model was carried out 9 times while the VGG-16 model was carried out 6 times as shown in Table 2. Based on a comprehensive and holistic analysis of the CNN model and VGG-16 model trials in Table 2, it can be explained that from a total of 14,085 image datasets with 80% train data and 20% test data, the best results were obtained in the eighth experiment, namely using the model test CNN with an epoch value of 50 and a learning rate value of 0.001, namely 0.9487 or 94.87%. The results of the data training can be seen in Figure 22. The algorithm model that was built succeeded in identifying pornographic and non-pornographic images with a confidence level of >75%.

Table 2. Comparison results of image content identification with CNN and VGG-16 models

| Model | Accuracy | Val accuracy | Losses | Val loss | Epoch | Learning rate |
|---|---|---|---|---|---|---|
| CNN 1 Trial 1 | 0.8341 | 0.8299 | 0.3691 | 0.2891 | 25 | 0.001 |
| Trial 2 | 0.8228 | 0.8038 | 0.3860 | 0.4707 | 25 | 0.0001 |
| Trial 3 | 0.8744 | 0.8289 | 0.2838 | 0.2281 | 50 | 0.001 |
| Trial 4 | 0.8503 | 0.7509 | 0.3475 | 0.2809 | 25 | 0.0001 |
| Trial 5 | 0.9066 | 0.8435 | 0.2108 | 0.2342 | 50 | 0.001 |
| Trial 6 | 0.8741 | 0.7673 | 0.3001 | 0.4083 | 50 | 0.0001 |
| Trial 7 | 0.8331 | 0.8254 | 0.3752 | 0.6289 | 25 | 0.0001 |
| Trial 8 | 0.9487 | 0.8831 | 0.1787 | 0.6308 | 50 | 0.001 |
| Trial 9 | 0.8122 | 0.7914 | 0.4170 | 0.5610 | 25 | 0.001 |
| VGG-16 Trial 1 | 0.5347 | 0.4375 | 0.6909 | 0.7083 | 100 | 0.001 |
| Trial 2 | 0.5198 | 0.4844 | 7.7400 | 0.7400 | 50 | 0.001 |
| Trial 3 | 0.4512 | 0.5188 | 8.8448 | 7.5554 | 50 | 0.001 |
| Trial 4 | 0.4674 | 0.5437 | 1.3211 | 3.5258 | 50 | 0.0001 |
| Trial 5 | 0.5344 | 0.5063 | 0.6908 | 0.6994 | 50 | 0.0001 |
| Trial 6 | 0.9531 | 0.8125 | 0.1214 | 0.3222 | 30 | 0.0001 |

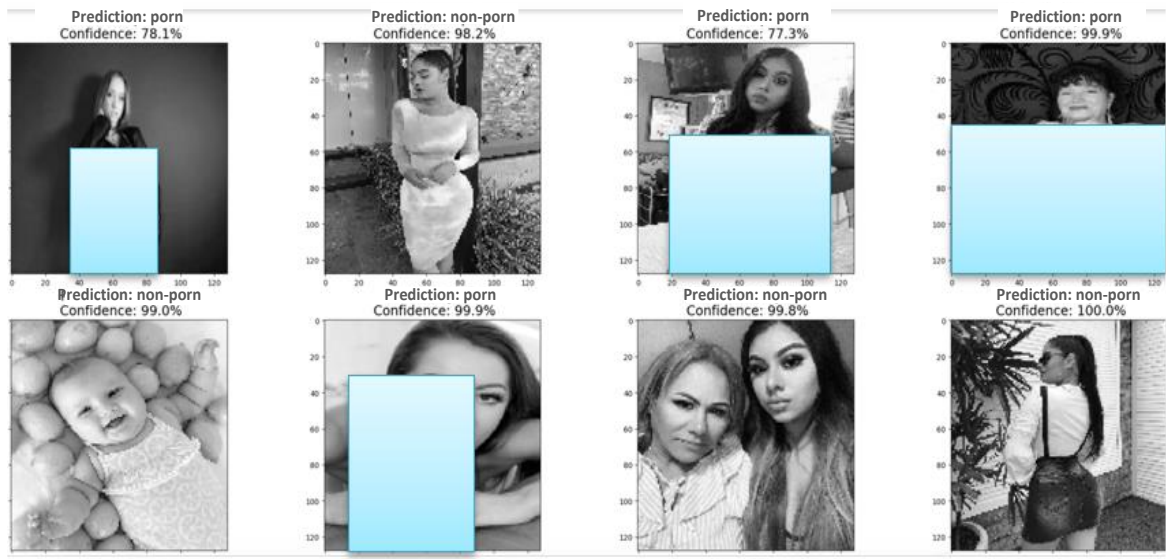

Figure 22. Image identification results with the CNN algorithm model

After obtaining the best model, an application prototype was created to make it easier for users to identify content that is indicated as pornography (Figure 23). This prototype is named SIKAPWEB. Through SIKAPWEB, users can input images to identify whether the image contains pornographic content or not.

During this research, we have extensively reviewed and evaluated the use of parameters in both models. The results of this trial help us determine the best combination of parameter characteristics to produce the optimal level of performance for each model. After analyzing comprehensively and holistically, we have now reached the final stage, CNN model with certain parameter settings produces maximum performance compared to the VGG-16 model. The most optimal level of accuracy was achieved in the CNN model (in the eighth experiment) by setting certain parameters which is a valuable contribution and can become a standard of results for future researchers, especially in the scope of deep learning theory.

The VGG-16 model is recognized for its reliability use small convolutional filters [31]. Thus, combining this model could increase the model's ability to capture complex features in detecting image content that is indicative of pornography while complementing the power of the CNN model [32]. VGG-16 can be a CNN model with its ability to process 16 layers. Now VGG-16 as a development of CNN has become the best computer vision model. This model is a form of network evaluation as well as maximizing the depth level by using architecture and very small convolution filters which produces a significant improvement compared to the previous configuration. VGG-16 model is the basic architecture for analyzing the spatial forms of image data [33].

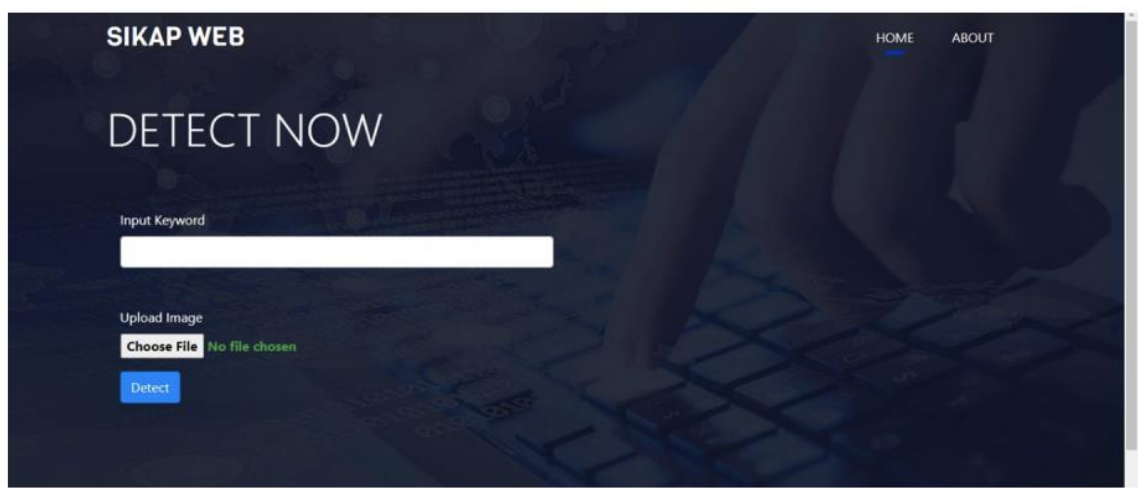

Figure 23. Prototype

This architecture allows the CNN model to effectively capture spatial relationships and patterns in images compared to the VGG-16 model. CNN have become popular for this purpose. CNN use multiple layers to model features and are applicable for image classification, object detection and image segmentation [34], [35]. Álvarez *et al*. [36] stated that the use of the CNN model network architecture was superior to other models in detecting negative and sensitive content such as image content containing pornographic elements. The CNN model in this research was built and trained using a data set that has been adjusted to the definition of pornography that applies in Indonesia and has obtained the best accuracy rate of 0.9487 or 94.87% in detecting images that fall into the pornographic content category.

There are 15 tests to evaluate which model and hyperparameters are the best in identifying pornographic content with 14,085 image data consisting of 9 CNN model tests using 1 layer, 2 layers, and 3 convolution layers and 6 VGG-16 model tests. Each test was carried out with different epochs and learning rates. The use of deep learning technology, especially CNN models, has shown significant potential in image recognition and classification. This holds great promise for more accurate and efficient automatic identification of pornographic content. With a large enough amount of image data (14,085 images), the model can be better trained to recognize patterns and features that indicate pornographic content.

Even though the amount of data is large, getting representative and high-quality data is still a challenge. The challenges faced in this research are that pornographic content has a wide variety and is often ambiguous because there are several images that are difficult to classify, whether they are pornographic or not, making classification difficult, the web scraping process requires a lot of time and resources and limited internet access. slow because you have to use VPN. Then the implementation of CNN and VGG-16 model requires adequate technological infrastructure for the computing process and data training.

The impact of the results of this research is that using the CNN and VGG-16 models can significantly increase accuracy in identifying pornographic image content. The implementation of CNN and VGG-16 model enables a faster and more efficient image detection process and enables scalability in monitoring and supervising content on the internet for the government of the Republic of Indonesia. Suggestions for future researchers are that more and varied datasets are needed to help the model learn better and increase its accuracy, as well as integrating the detection model with other monitoring systems such as text, audio, and metadata to provide a more complete context and increase the accuracy of detecting pornographic content.

## 5. CONCLUSION

The results of the comprehensive and holistic analysis study in the previous section regarding the comparison of the test results of the CNN model and the approach with the VGG-16 model, it can be explained that from a total of 14,085 image datasets with 80% train data and 20% test data, the best results were obtained in the eighth experiment, namely using a CNN model test with an epoch value of 50 and a learning rate value of 0.001, namely 0.9487 or 94.87%. To obtain comprehensive and holistic research results and overcome research limitations and project the impact of the latest research results, researchers can further develop other objects with higher data quality and better deep learning algorithm model architecture with a higher level of accuracy and data processing performance which is faster. It is hoped that future perspectives from these findings can be deepened and expanded sustainably with extensive data and reproduced as plugins, especially as child-friendly internet filters. It is important to understand that studies on image content classification using deep learning models will continue to develop in the future and this research cannot generalize to all data in the field. However, the results of this research can provide a high benchmark for further research and future development of informatics management science, especially new research on deep learning. Further research developments could include modifications to the CNN model to further optimize elements of precision and accuracy in image content classification. Holistically, the results of this research are expected to provide valuable knowledge contribution to the development of image content classification theory and become a basis for future research.

## ACKNOWLEDGEMENTS

On this occasion, the author would like to express deepest gratitude to the Yayasan Pendidikan Gunadarma for providing a valuable opportunity to continue studies in the Doctoral Program in Information Technology at Gunadarma University. Prof. Dr. E.S. Margianti, SE., MM., Rector of Gunadarma University and Prof. Suryadi Harmanto, SSi., MMSI., Vice Rector II of Gunadarma University, for providing the opportunity and trust to continue these studies with all the facilities, support, encouragement, and conveniences given, enabling the coursework, research, and dissertation writing processes to proceed smoothly.

## FUNDING INFORMATION

Authors state no funding involved. This research was independently funded by the authors.

## AUTHOR CONTRIBUTIONS STATEMENT

This journal uses the Contributor Roles Taxonomy (CRediT) to recognize individual author contributions, reduce authorship disputes, and facilitate collaboration

| Name of Author | C | M | So | Va | Fo | I | R | D | O | E | Vi | Su | P | Fu |
|---|---|---|---|---|---|---|---|---|---|---|---|---|---|---|
| Reza Chandra | ✓ | ✓ | ✓ | ✓ | ✓ | ✓ | ✓ |  | ✓ | ✓ | ✓ |  | ✓ |  |
| Adang Suhendra | ✓ | ✓ |  | ✓ | ✓ | ✓ | ✓ |  |  | ✓ |  | ✓ |  |  |
| Lintang Yuniar Banowosari |  | ✓ |  | ✓ |  | ✓ |  | ✓ | ✓ | ✓ |  | ✓ | ✓ |  |
| Prihandoko |  | ✓ |  | ✓ |  | ✓ |  | ✓ |  | ✓ | ✓ |  |  |  |

| C  : **C**onceptualization | I  : **I**nvestigation | Vi : **Vi**sualization |
|---|---|---|
| M  : **M**ethodology | R  : **R**esources | Su : **Su**pervision |
| So : **So**ftware | D  : **D**ata Curation | P  : **P**roject administration |
| Va : **Va**lidation | O  : Writing - **O**riginal Draft | Fu : **Fu**nding acquisition |
| Fo : **Fo**rmal analysis | E  : Writing - Review & **E**diting | |

## CONFLICT OF INTEREST STATEMENT

Authors state no conflict of interest.

## INFORMED CONSENT

We have obtained informed consent from all individuals included in this study.

## ETHICAL APPROVAL

The research related to human use has been complied with all the relevant national regulations and institutional policies in accordance with the tenets of the Helsinki Declaration and has been approved by the authors' institutional review board or equivalent committee.

# DATA AVAILABILITY

The data that support the findings of this study are available on request from the author, [RC]. The data, which contain information that could compromise the privacy of research participants, are not publicly available due to certain restrictions.

# REFERENCES

[1]   S. Siswanto, "Pancasila as strategy to prevent proxy war," *Jurnal Pertahanan*, vol. 3, no. 2, 2017, doi: 10.33172/jp.v3i2.218.
[2]   N. Ulfah, N. O. Irawan, P. D. Nurfadila, P. Y. Ristanti, and J. A. Hammad, "Blocking pornography sites on the internet private and university access," *Bulletin of Social Informatics Theory and Application*, vol. 3, no. 1, pp. 22–29, 2019, doi: 10.31763/businta.v3i1.161.
[3]   N. Thurman and F. Obster, "The regulation of internet pornography: What a survey of under-18s tells us about the necessity for and potential efficacy of emerging legislative approaches," *Policy & Internet*, vol. 13, no. 3, pp. 415–432, 2021, doi: 10.1002/poi3.250.
[4]   M. Andriansyah, I. Purwanto, M. Subali, A. I. Sukowati, M. Samos, and A. Akbar, "Developing Indonesian corpus of pornography using simple NLP-text mining (NTM) approach to support government anti-pornography program," in *2017 Second International Conference on Informatics and Computing (ICIC)*, Nov. 2017, pp. 1–4. doi: 10.1109/IAC.2017.8280618.
[5]   C. Keen, R. Kramer, and A. France, "The pornographic state: the changing nature of state regulation in addressing illegal and harmful online content," *Media, Culture & Society*, vol. 42, no. 7–8, pp. 1175–1192, Oct. 2020, doi: 10.1177/0163443720904631.
[6]   R. Ballester-Arnal, M. García-Barba, J. Castro-Calvo, C. Giménez-García, and M. D. Gil-Llario, "Pornography Consumption in people of different age groups: an analysis based on gender, contents, and consequences," *Sexuality Research and Social Policy*, vol. 20, no. 2, pp. 766–779, Jun. 2023, doi: 10.1007/s13178-022-00720-z.
[7]   D. P. Fernandez, D. J. Kuss, L. V. Justice, E. F. Fernandez, and M. D. Griffiths, "Effect of a 7-day pornography abstinence period on withdrawal-related symptoms in regular pornography users: a randomized controlled study," *Archives of Sexual Behavior*, vol. 52, no. 4, pp. 1819–1840, 2023, doi: 10.1007/s10508-022-02519-w.
[8]   W. Hu, O. Wu, Z. Chen, Z. Fu, and S. Maybank, "Recognition of pornographic web pages by classifying texts and images," *IEEE Transactions on Pattern Analysis and Machine Intelligence*, vol. 29, no. 6, pp. 1019–1034, Jun. 2007, doi: 10.1109/TPAMI.2007.1133.
[9]   Y. Zhang and B. Wallace, "A sensitivity analysis of (and practitioners' guide to) convolutional neural networks for sentence classification," *arXiv-Computer Science*, pp. 1-18, Oct. 2015.
[10]  J. Patterson and A. Gibson, *Deep learning: a practitioner's approach*, 1st ed. Sebastopol, California: O'Reilly Media, 2017.
[11]  L. Alzubaidi *et al.*, "Review of deep learning: concepts, CNN architectures, challenges, applications, future directions," *Journal of Big Data*, vol. 8, no. 1, 2021, doi: 10.1186/s40537-021-00444-8.
[12]  L. Chen, S. Li, Q. Bai, J. Yang, S. Jiang, and Y. Miao, "Review of image classification algorithms based on convolutional neural networks," *Remote Sensing*, vol. 13, no. 22, 2021, doi: 10.3390/rs13224712.
[13]  I. M. A. Agastya, A. Setyanto, Kusrini, and D. O. D. Handayani, "Convolutional Neural Network for Pornographic Images Classification," in *2018 Fourth International Conference on Advances in Computing, Communication & Automation (ICACCA)*, pp. 1–5, 2018, doi: 10.1109/ICACCAF.2018.8776843.
[14]  R. M. Alguliyev, F. J. Abdullayeva, and S. S. Ojagverdiyeva, "Image-based malicious Internet content filtering method for child protection," *Journal of Information Security and Applications*, vol. 65, Mar. 2022, doi: 10.1016/j.jisa.2022.103123.
[15]  A. Pandey, S. Moharana, D. P. Mohanty, A. Panwar, D. Agarwal, and S. P. Thota, "On-device content moderation," *arXiv-Computer Science*, pp. 1-7, 2021.
[16]  C. Chakraborty, M. Bhattacharya, S. Pal, and S.-S. Lee, "From machine learning to deep learning: Advances of the recent data-driven paradigm shift in medicine and healthcare," *Current Research in Biotechnology*, vol. 7, pp. 1–20, 2024, doi: 10.1016/j.crbiot.2023.100164.
[17]  I. H. Sarker, "Deep learning: a comprehensive overview on techniques, taxonomy, applications and research directions," *SN Computer Science*, vol. 2, no. 6, Nov. 2021, doi: 10.1007/s42979-021-00815-1.
[18]  J. Kufel *et al.*, "What is machine learning, artificial neural networks and deep learning?—examples of practical applications in medicine," *Diagnostics*, vol. 13, no. 15, Aug. 2023, doi: 10.3390/diagnostics13152582.
[19]  M. M. Taye, "Understanding of machine learning with deep learning: architectures, workflow, applications and future directions," *Computers*, vol. 12, no. 5, Apr. 2023, doi: 10.3390/computers12050091.
[20]  I. Goodfellow, Y. Bengio, and A. Courville, *Deep learning*. Cambridge, Massachusetts: MIT Press, 2016.
[21]  D. Bhatt *et al.*, "CNN variants for computer vision: history, architecture, application, challenges and future scope," *Electronics*, vol. 10, no. 20, Oct. 2021, doi: 10.3390/electronics10202470.
[22]  T. Karlita, N. A. Choirunisa, R. Asmara, and F. Setyorini, "Cat breeds classification using compound model scaling convolutional neural networks," in *International Conference on Applied Science and Technology on Social Science 2021 (iCAST-SS 2021)*, 2022, doi: 10.2991/assehr.k.220301.150.
[23]  X. Zhao, J. Ma, L. Wang, Z. Zhang, Y. Ding, and X. Xiao, "A review of hyperspectral image classification based on graph neural networks," *Artificial Intelligence Review*, vol. 58, no. 6, Mar. 2025, doi: 10.1007/s10462-025-11169-y.
[24]  M. J. Sudhamani, I. Sanyal, and M. K. Venkatesha, "Artificial neural network approach for multimodal biometric authentication system," in *Proceedings of Data Analytics and Management*, Springer, pp. 253–265, 2022, doi: 10.1007/978-981-16-6285-0_21.
[25]  Y. Wang, D. Shi, and W. Zhou, "Convolutional neural network approach based on multimodal biometric system with fusion of face and finger vein features," *Sensors*, vol. 22, no. 16, Aug. 2022, doi: 10.3390/s22166039.
[26]  Y. Yin *et al.*, "Artificial neural networks for finger vein recognition: a survey," *Engineering Applications of Artificial Intelligence*, vol. 150, 2025, doi: 10.1016/j.engappai.2025.110586.
[27]  S. M. Kadhim, J. K. S. Paw, Y. C. Tak, and S. Ameen, "Deep learning models for biometric recognition based on face, finger vein, fingerprint and iris: a survey," *Journal of Smart Internet of Things*, vol. 2024, no. 1, pp. 117–157, 2024, doi: 10.2478/jsiot-2024-0007.
[28]  F. D. Adhinata, N. A. F. Tanjung, W. Widayat, G. R. Pasfica, and F. R. Satura, "Comparative study of VGG16 and MobileNetV2 for masked face recognition," *Jurnal Ilmiah Teknik Elektro Komputer dan Informatika*, vol. 7, no. 2, 2021, doi: 10.26555/jiteki.v7i2.20758.
[29]  Q. Guan *et al.*, "Deep convolutional neural network VGG-16 model for differential diagnosing of papillary thyroid carcinomas in cytological images: a pilot study," *Journal of Cancer*, vol. 10, no. 20, pp. 4876–4882, 2019, doi: 10.7150/jca.28769.
[30]  A. Coates, A. Ng, and H. Lee, "An analysis of single-layer networks in unsupervised feature learning," in *Proceedings of the Fourteenth International Conference on Artificial Intelligence and Statistics*, pp. 215–223, 2011.

[31] M. Jain, M. S. Bora, S. Chandnani, A. Grover, and S. Sadwal, "Comparison of VGG-16, VGG-19, and ResNet-101 CNN models for the purpose of suspicious activity detection," *International Journal of Scientific Research in Computer Science, Engineering and Information Technology*, vol. 9, no. 1, pp. 121–130, Jan. 2023, doi: 10.32628/CSEIT2390124.
[32] S. B. Boddu, A. V. Kanumuri, D. C. R. Triveni, G. V. S. S. Prasanth, and S. R. Reddy, "Fake images detection: a comparative study using CNN and VGG-16 models," *International Journal of Advances in Engineering and Management*, vol. 5, no. 11, pp. 150–157, 2023, doi: 10.35629/5252-0511150157.
[33] W. Al-Khater and S. Al-Madeed, "Using 3D-VGG-16 and 3D-Resnet-18 deep learning models and FABEMD techniques in the detection of malware," *Alexandria Engineering Journal*, vol. 89, pp. 39–52, Feb. 2024, doi: 10.1016/j.aej.2023.12.061.
[34] M. Krichen, "Convolutional neural networks: a survey," *Computers*, vol. 12, no. 8, 2023, doi: 10.3390/computers12080151.
[35] C.-L. Fan, "Deep neural networks for automated damage classification in image-based visual data of reinforced concrete structures," *Heliyon*, vol. 10, no. 19, 2024, doi: 10.1016/j.heliyon.2024.e38104.
[36] D. P. Álvarez, A. L. S. Orozco, J. P. García-Miguel, and L. J. G. Villalba, "Learning strategies for sensitive content detection," *Electronics*, vol. 12, no. 11, Jun. 2023, doi: 10.3390/electronics12112496.

## BIOGRAPHIES OF AUTHORS

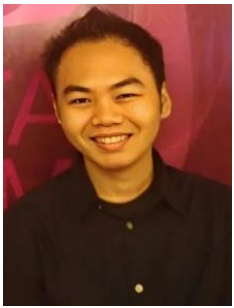
**Reza Chandra** holds a Doctor of Information Technology from Gunadarma University, Indonesia in 2020. He also received his Bachelor Degree (information systems) and Master Degree (information systems management) from Gunadarma University, Indonesia in 2009 and 2012, respectively. He is currently a lecturer at Department of Informatics Management in Gunadarma University, Indonesia. His research includes information systems, machine learning, data mining, deep learning, and computer networks. He has published over 13 papers in international journals and conferences. He can be contacted at email: reza_chan@staff.gunadarma.ac.id.

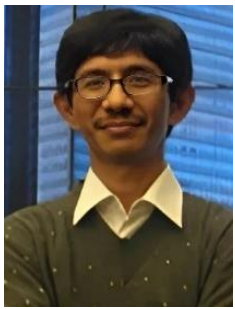
**Adang Suhendra** holds a Doktoringenieur from Universitat Kassel, Germany in 2001. He also received his Bachelor Degree in information systems from Gunadarma University and Bachelor Degree in mathematics from University of Indonesia in 1992 and 1993 and Master Degree (Master of Science) from Asian Institute of Technology, Thailand in 1994. He is currently Professor in information technology at Gunadarma University. His research includes mathematical computing, machine learning, data mining, deep learning, and big data. He has published over 90 papers in international journals and conferences. He can be contacted at email: adang@staff.gunadarma.ac.id.

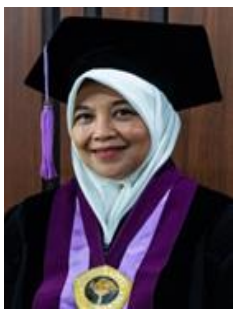
**Lintang Yuniar Banowosari** holds a Doctor of Information Technology from Gunadarma University, Indonesia in 2007. She also received her Bachelor Degree (information systems) Indonesia in 1993 and Master Degree (Master of Science) from Asian Institute of Technology, Thailand in 1994. She is currently Professor in information technology at Gunadarma University. Her research includes information systems, machine learning, data mining, deep learning, and ontology. She has published over 190 papers in international journals and conferences. She can be contacted at email: lintang@staff.gunadarma.ac.id.

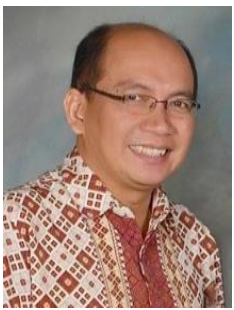
**Prihandoko** holds a Doctor of Computer Communication from Universiti Putra Malaysia in 2003. He also received his Bachelor Degree (computer science) from the University of Indonesia in 1991 and Master Degree (information technology) from The University of Queensland, Australia, in 1998. He is currently Associate Professor at Department of Information Systems in Gunadarma University, Indonesia. His research includes database design, software engineering, image processing, mobile computing, and information systems. He has published over 70 papers in international journals and conferences. He can be contacted at email: pri@staff.gunadarma.ac.id.